\pdfoutput=1

\documentclass[10pt,twocolumn,letterpaper]{article}

\usepackage[pagenumbers]{cvpr} % To force page numbers, e.g. for an arXiv version

\usepackage{graphicx}
\usepackage{amsmath}
\usepackage{amssymb}
\usepackage{booktabs}
\usepackage{multirow}
\usepackage{bbm}
\usepackage{algorithm,algpseudocode}
\usepackage{wrapfig}

\usepackage[pagebackref,breaklinks,colorlinks]{hyperref}

\usepackage[capitalize]{cleveref}
\crefname{section}{Sec.}{Secs.}
\Crefname{section}{Section}{Sections}
\Crefname{table}{Table}{Tables}
\crefname{table}{Tab.}{Tabs.}

\def\cvprPaperID{107} % *** Enter the CVPR Paper ID here
\def\confName{CVPR}
\def\confYear{2023}

\begin{document}

% \title{Not All Paths and Data Are Equal for One-Shot Neural Architecture Search}
% \title{PDD: Importance-based Paths and Data Dual Sampling for One-shot Neural Architecture Search\\
% \jianchao{PDD: Importance-based Paths and Data Dual sampling for one-shot neural architecture search} \\
% \jianchao{PDD: Paths and Data Dual sampling for one-shot neural architecture search}
% }
\title{PA\&DA: Jointly Sampling PAth and DAta for Consistent NAS}

\author{
% Authors
Shun Lu\textsuperscript{\rm 1, \rm 2},  
Yu Hu\textsuperscript{\rm 1, \rm 2}\thanks{Corresponding author.}, 
Longxing Yang\textsuperscript{\rm 1, \rm 2}, 
Zihao Sun\textsuperscript{\rm 1, \rm 2}, 
Jilin Mei\textsuperscript{\rm 1}, 
Jianchao Tan\textsuperscript{\rm 3}, 
Chengru Song\textsuperscript{\rm 3} \\
% Institutions 
\textsuperscript{\rm 1} Research Center for Intelligent Computing Systems, \\Institute of Computing Technology, Chinese Academy of Sciences \\
    \textsuperscript{\rm 2} School of Computer Science and
Technology, University of Chinese Academy of Sciences \\
    \textsuperscript{\rm 3} Kuaishou Technology \\
% Emails 
{\tt\small \{lushun19s, huyu, yanglongxing20b, sunzihao18z, meijilin\}@ict.ac.cn,}\\
{\tt\small \{jianchaotan, songchengru\}@kuaishou.com}
}

%% omit the following lines up until the closing ``}''.
%% Additional authors and addresses can be added with ``\and'',
%% just like the second author.
%% To save space, use either the email address or home page, not both
%\and
\maketitle

%%%%%%%%% ABSTRACT
\begin{abstract}
%   Based on the weight-sharing mechanism, one-shot neural architecture search (NAS) methods train a supernet and then inherit the pre-trained supernet weights to evaluate the performance of sub-models, liberating conventional NAS from the unaffordable computational burden. 
%   However, we observe that the weight-sharing mechanism inevitably increases the supernet gradient variance during training, degrading the supernet performance and leading to a poor ranking consistency of sub-models. 

% Short version
  Based on the weight-sharing mechanism, one-shot NAS methods train a supernet and then inherit the pre-trained weights to evaluate sub-models, largely reducing the search cost. 
  However, several works have pointed out that the shared weights suffer from different gradient descent directions during training. And we further find that large gradient variance occurs during supernet training, which degrades the supernet ranking consistency. 
  To mitigate this issue, we propose to explicitly minimize the gradient variance of the supernet training by jointly optimizing the sampling distributions of \textbf{PA}th and \textbf{DA}ta (PA\&DA).
  We theoretically derive the relationship between the gradient variance and the sampling distributions, and reveal that the optimal sampling probability is proportional to the normalized gradient norm of path and training data. 
  Hence, we use the normalized gradient norm as the importance indicator for path and training data, and adopt an importance sampling strategy for the supernet training.
  Our method only requires negligible computation cost for optimizing the sampling distributions of path and data, but achieves lower gradient variance during supernet training and better generalization performance for the supernet, resulting in a more consistent NAS.
  We conduct comprehensive comparisons with other improved approaches in various search spaces. 
  Results show that our method surpasses others with more reliable ranking performance and higher accuracy of searched architectures, showing the effectiveness of our method. 
  Code is available at \href{https://github.com/ShunLu91/PA-DA}{https://github.com/ShunLu91/PA-DA}.

\end{abstract}

%%%%%%%%% BODY TEXT
\section{Introduction}
\label{sec:intro}

Neural architecture search (NAS) aims to automate the process of designing architectures. 
Conventional NAS methods \cite{baker2017designing, zoph2017neural} separately train each sub-model to guide the controller for better architectures, thus demanding prohibitive computational complexity. 
ENAS \cite{pham2018efficient} explores the weight-sharing mechanism across sub-models and One-Shot NAS \cite{bender2018understanding} further proposes to train a supernet to share the pre-trained weights for sub-models to achieve higher efficiency. 
Later on, many follow-ups \cite{liu2019darts, xie2019snas, guo2020single, chu2021fairnas} adopt this vein to perform NAS.

Though the weight-sharing mechanism greatly improves NAS efficiency, many works \cite{zhang2020overcoming, zhao2021few, su2021k, hu2022generalizing, zhou2022close, ha2021sumnas, xu2022analyzing} point out that the shared weights suffer from different gradient descent directions in different sub-models, leading to large gradient variance and poor ranking consistency. 
To mitigate this issue, they \cite{zhao2021few, su2021k, hu2022generalizing, zhou2022close} propose to maintain multi-copies of supernet weights to decrease the weight-sharing extent, manually elaborate a better path sampling strategy \cite{chu2021fairnas, xu2022analyzing}, or introduce additional loss regularizations \cite{zhang2020overcoming, ha2021sumnas, xu2022analyzing}.
However, they typically require multiple computation burdens for the supernet training and obtain unsatisfying results, motivating us to explore a better solution. 

Notice that significant efforts \cite{roux2012stochastic, shalev2012stochastic, johnson2013accelerating, defazio2014saga, schmidt2017minimizing, gorbunov2020unified} have been dedicated to reducing the variance of the stochastic gradient descent (SGD) for minimizing finite sums. 
And many works \cite{zhao2015stochastic, alain2015variance, gopal2016adaptive, chang2017active, wang2017accelerating, johnson2018training, katharopoulos2018not} focus on optimizing the data sampling distribution to reduce the gradient variance for training deep models. 
These methods generally enjoy faster convergence and better generalization performance, inspiring us to improve the supernet training from the perspective of gradient variance reduction.

\begin{figure}[t]
  \centering
  \subcaptionbox{Different weight-sharing extent}{
  \includegraphics[width=0.48\linewidth]{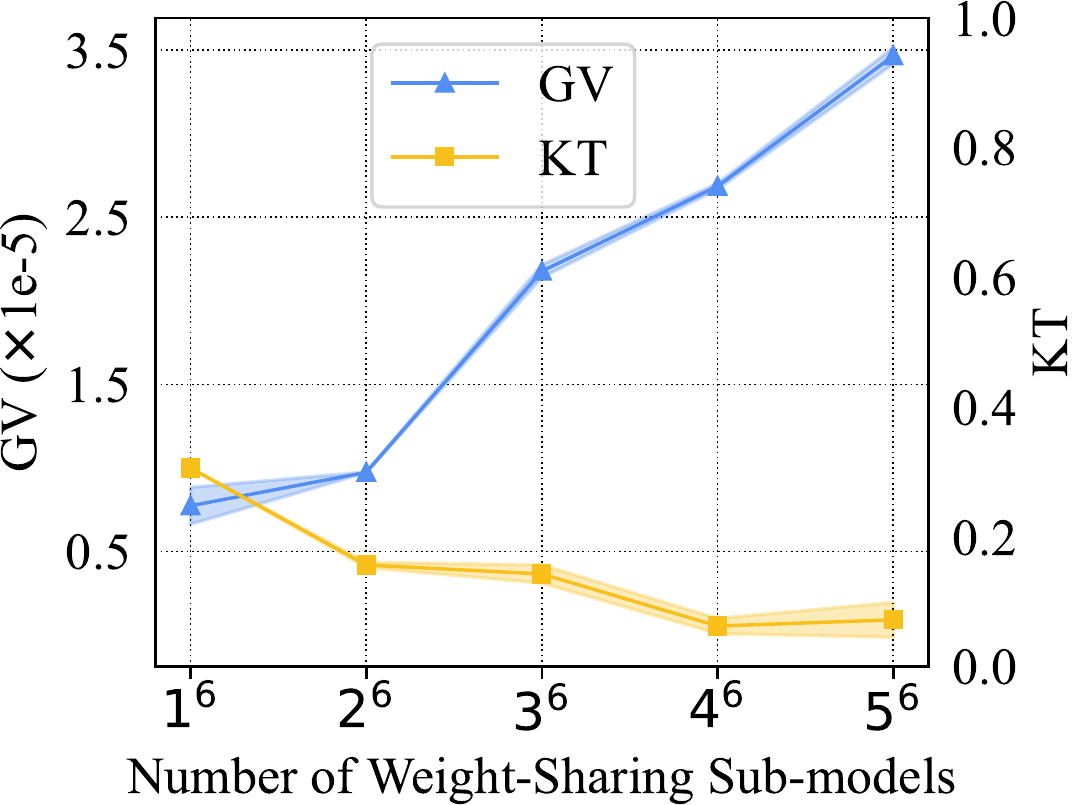}}
  \subcaptionbox{Different methods}{
  \includegraphics[width=0.48\linewidth]{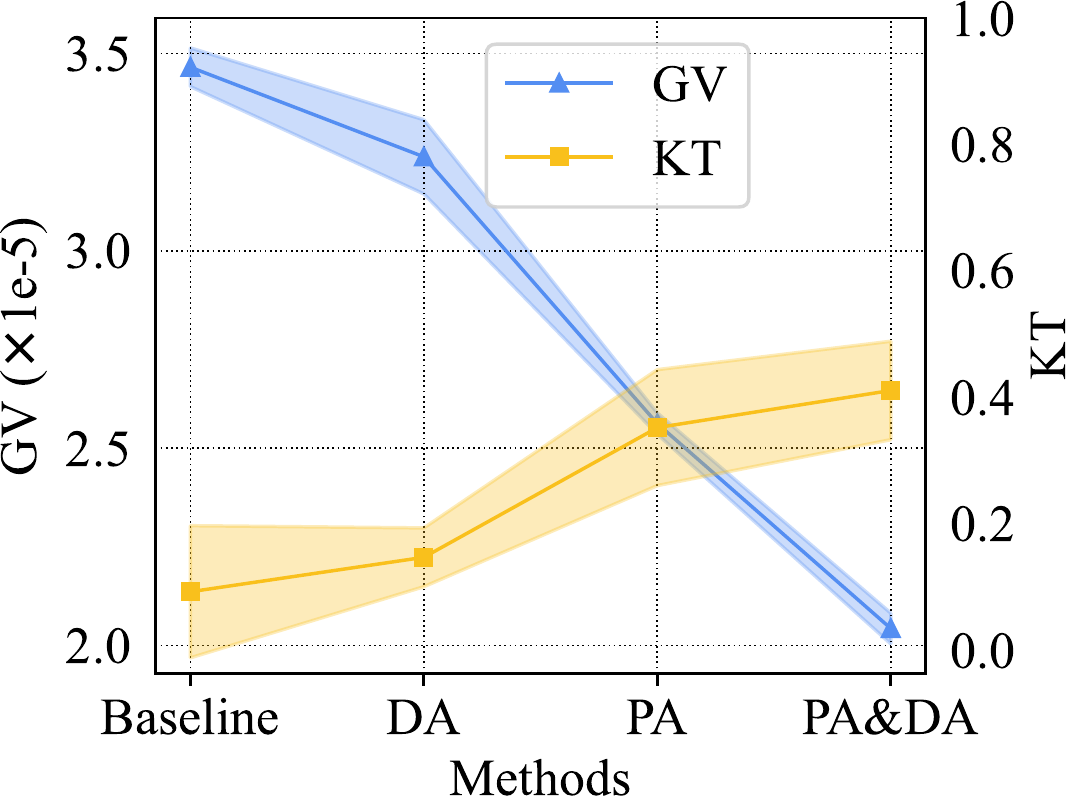}}
  \caption{Left: The trend of KT and GV when we change the weight-sharing extent by increasing the number of weight-sharing sub-models exponentially. Right: The comparison of KT and GV between the baseline and our method. KT: Kendall's Tau, GV: Gradient Variance. Detailed GV calculation is in the supplements.}
  \label{fig:observation}
  \vspace{-2mm}
\end{figure}

We conduct a toy experiment on NAS-Bench-201 \cite{dong2020bench} using CIFAR-10 to investigate the gradient variance for a weight-sharing supernet during training. 
We adopt the SPOS \cite{guo2020single} algorithm to train the supernet and gradually increase the candidate operations on each edge of the supernet to change the weight-sharing extent. 
We record the average gradient variance of all candidate operation weights during training and evaluate the supernet performance by measuring the ranking results of the same 64 sub-models (corresponding to the smallest search space with 2 candidate operations on 6 edges). 
As illustrated in Fig.\ref{fig:observation}(a), with more sub-models sharing weights from the supernet, the gradient variance becomes larger and the ranking consistency becomes worse.
These results indicate that a larger gradient variance during training harms the supernet ranking consistency, which prompts us to reduce the gradient variance to improve the supernet performance.

In this paper, we aim to reduce the supernet gradient variance during training to improve the convergence rate and generalization performance. 
We propose to jointly optimize the path and data sampling distributions during supernet training to achieve this goal. 
We theoretically derive the relationship between the supernet gradient variance and the sampling distributions and then explicitly minimize the gradient variance by optimizing the path and data sampling distributions respectively. 
We find that the optimal sampling probability is proportional to the normalized gradient norm of the path and training data. 
Thus we use the normalized gradient norm as the importance indicator and adopt an importance sampling strategy for path and data during the supernet training.
As exemplified in Fig.\ref{fig:observation}(b), by using our proposed PAth importance sampling (PA) and DAta importance sampling (DA), we can reduce the supernet gradient variance and improve its ranking consistency.

Overall, our contributions are as follows:
\begin{itemize}
    \item We validate that the weight-sharing mechanism for the supernet training induces large gradient variance, harming the supernet performance and worsening its ranking consistency.
    \item By deriving the relationship between the supernet gradient variance and sampling distributions, we propose to explicitly minimize the gradient variance by \textbf{jointly optimizing path and data sampling distributions} during supernet training. We reveal that the optimal sampling probability is proportional to the normalized gradient norm of path and data, and adopt an importance sampling for them during the supernet training.
    \item Our method only requires negligible computation to perform the importance sampling for path and data, and does not need tediously hyper-parameter tuning. We obtain the highest Kendall's Tau \cite{sen1968estimates} 0.713 on NAS-Bench-201 and achieve superior performance on DARTS and ProxylessNAS search spaces.
\end{itemize}

\section{Related Work}
\label{sec:related_work}

\subsection{One-Shot Neural Architecture Search}
Early NAS methods \cite{baker2017designing, zoph2017neural} are time-consuming due to the training of each sub-model from scratch. ENAS \cite{pham2018efficient} leverages the weight-sharing mechanism to share weights across sub-models, greatly speeding up the NAS process.
One-Shot NAS \cite{bender2018understanding} proposes to train the supernet using the path dropout technique and then conduct the architecture search by inheriting the pre-trained supernet weights to sub-models for fast evaluation. 
Follow-up works can be roughly divided into gradient-based approaches \cite{liu2019darts, xie2019snas} and sampling-based methods \cite{guo2020single, chu2020fair}. 
Gradient-based approaches relax the architectural parameters to be continuous and jointly optimize the architectural parameters and supernet weights using the gradient descent, which requires multiple memory costs and often suffers from instability. 
While sampling-based methods train the supernet weights by sampling different paths (i.e. sub-models) and only optimize one path at each training step, which is more efficient and robust in practice. 
For example, the widely used SPOS \cite{guo2020single} algorithm adopts a uniform sampling strategy to sample path and training data during the supernet training.
In this work, we focus on the sampling-based one-shot NAS methods and utilize SPOS as our baseline.

\subsection{Improved Sampling-based One-Shot NAS}
To enhance the consistency of one-shot NAS, some works \cite{zhao2021few, su2021k, hu2022generalizing, zhou2022close} maintain multiple copies of supernet weights to decrease the weight-sharing extent. 
They focus on how to split the supernet and how to select appropriate weights for a sampled sub-model, again complicating the supernet training. 
Several works try to seek a more reasonable gradient direction for better convergence. 
NSAS \cite{zhang2020overcoming} utilizes the loss regularization to prevent the performance of other sub-models from degrading and SUMNAS \cite{ha2021sumnas} computes the reptile gradient during supernet training. Both of them are motivated by multi-model forgetting and consume multiple computations during training. 
Other works manually elaborate a better path sampling strategy to handle this issue. 
FairNAS \cite{chu2020fair} samples and trains candidate operations without replacement and accumulate the gradients until all of them are activated, ensuring strict fairness to benefit the supernet training. 
MAGIC-AT \cite{xu2022analyzing} increases the gradient similarity between sampled architectures by substituting only one candidate operation across consecutively sampled paths and employing an alignment loss for supernet training. 
They require lots of human experience and more computation resources. 
In this work, we optimize the sampling distributions of path and data according to the normalized gradient norm, thus maintaining the efficiency and improving the consistency of one-shot NAS without the need for complex manual design. 

\subsection{Variance Reduction}
Many approaches \cite{roux2012stochastic, shalev2012stochastic, johnson2013accelerating, defazio2014saga, schmidt2017minimizing, gorbunov2020unified} achieve linear convergence on empirical risk minimization problems by reducing the variance of SGD. 
They enjoy faster convergence and better generalization performance than vanilla SGD. 
Other works \cite{zhao2015stochastic, alain2015variance, gopal2016adaptive, chang2017active, wang2017accelerating, johnson2018training, katharopoulos2018not} optimize the data sampling distribution during training to reduce the stochastic gradient variance. 
They \cite{needell2014stochastic, zhao2015stochastic} derive a clear relationship that the optimal sampling distribution is proportional to the per-sample gradient norm and use an importance sampling \cite{katharopoulos2018not, johnson2018training, jiang2019accelerating, coleman2019selection} for the training data. 
As common deep learning frameworks only provide the average gradient of a mini-batch, it is prohibitively expensive to compute the per-sample gradient norm. To solve this problem, \cite{gopal2016adaptive} exploits the side information to model the sampling distribution per class instead of per sample and \cite{katharopoulos2018not} approximates the upper bound of the per-sample gradient norm efficiently. 
Our work is motivated by these methods. Differently, we focus on sampling different paths for supernet training using the normalized gradient norm, which can be efficiently acquired across mini-batches. 
Besides, we utilize the approximation from \cite{katharopoulos2018not} to perform an importance sampling for training data during the supernet optimization. 
% We also notice that DA-NAS \cite{dai2020data} have employed the curriculum learning method to sample the training data for NAS. They utilize a pre-trained model to sort the training data from easy to hard in advance and sequentially feed them into the supernet during training. We provide a fair comparison with DA-NAS in our experiments.

\section{Method}
\label{sec:method}

\subsection{Sampling-based One-Shot NAS}

One-Shot NAS \cite{pham2018efficient, bender2018understanding} has recently become mainstream due to its efficiency and simplicity. Particularly, sampling-based one-shot NAS approaches \cite{guo2020single, chu2021fairnas} demonstrate superior performance in searching for top-performing architectures. 
These methods generally have two stages, i.e., supernet training and sub-model searching. 

In the first stage, a supernet $\mathcal{N}$ with weights $\mathcal{W}$ is built by encoding the whole search space $\mathcal{A}$. 
During training, a sub-model $\alpha$ is sampled according to the discrete distribution $\textbf{p}(\mathcal{A})$ and we only train the weights $\mathcal{W_\alpha}$ included in the sampled sub-model at each step. 
We aim to obtain the optimal supernet weights $\mathcal{W}^\star$ by iteratively sampling and training the sampled sub-models with the training loss $\mathcal{L}$, 
\begin{equation}
\mathcal{W}^\star = \underset{\mathcal{W}}{\mathrm{argmin}} \ \mathbb{E}_{\substack{\alpha \sim \textbf{p}(\mathcal{A}) \\ (x,y) \sim \textbf{q}(\mathbb{D}_T)}}[\mathcal{L}(\mathcal{N}(x,\alpha; \mathcal{W_\alpha}), y)]
\label{eq:supernet_train_obj}
\end{equation}
% \begin{equation}
% \mathcal{W}^\star = \underset{\mathcal{W}}{\mathrm{argmin}} \ \mathbb{E}_{\substack{\alpha \widesim{p_{\alpha}} \Gamma(\mathcal{A}) \\ (x,y) \widesim{q_{(x,y)}} \Omega(\mathbb{D}_T)}}[\mathcal{L}(\mathcal{N}(x,\alpha; \mathcal{W_\alpha}), y)]
% \label{eq:supernet_train_obj}
% \end{equation}
where $(x, y)$ is sampled from the training dataset $\mathbb{D}_T$ according to the distribution $\textbf{q}(\mathbb{D}_T)$. 

In the second stage, we inherit the optimal supernet weights $\mathcal{W}^\star$ for each sub-model to efficiently evaluate their performance $\mathcal{P}$ on the validation dataset $\mathbb{D}_V$. 
A heuristic search algorithm is often applied to search for the top-performing sub-model $\alpha^\star$, 
\begin{equation}
\alpha^\star=\underset{\alpha \in \mathcal{A}}{\mathrm{argmax}} \ \mathbb{E}_{(x, y) \sim \textbf{q}(\mathbb{D}_V)}[\mathcal{P}(\mathcal{N}(x, \alpha; \mathcal{W^\star_\alpha}), y)]
\end{equation}

The performance of each sub-model is measured using the supernet weights, thus the supernet ranking consistency becomes essential for the ultimate NAS performance. 
We try to reduce the supernet gradient variance during training to improve the supernet convergence and ranking consistency. 
We propose to jointly optimize the sampling distributions of $\textbf{p}(\mathcal{A})$ and $\textbf{q}(\mathbb{D}_T)$ during the supernet training, which implies a bi-level optimization problem with $\mathcal{W}$ as the upper-level variable and $\textbf{p}, \textbf{q}$ as the lower-level variable
\begin{equation}
\begin{aligned}
& \quad \mathcal{W}^\star = \underset{\mathcal{W}}{\mathrm{argmin}} \ \mathbb{E}[\mathcal{L}(\mathcal{N}(x,\alpha; \mathcal{W_\alpha}), y)] \\
\mathrm{ s.t. } \ & 
\left\{\begin{array}{lr}
\alpha \sim \textbf{p}^\star(\mathcal{A}), \
(x, y) \sim \textbf{q}^\star(\mathbb{D}_T), \\\\
\textbf{p}^\star = \underset{\textbf{p}}{\mathrm{argmin}} \mathbb{V}[d(\textbf{p})], \\
\textbf{q}^\star = \underset{\textbf{q}}{\mathrm{argmin}} \mathbb{V}[d(\textbf{q})]
\end{array} \right.
\end{aligned}
\label{eq:ensemble_formulation}
\end{equation}
where $d(\textbf{p})$ and $d(\textbf{q})$ are the gradient variance function regarding the path and data sampling distributions. 
In the following, we introduce how to derive their relationship and alternatively optimize both sampling distributions.

\begin{figure*}[t]
  \centering
  \includegraphics[width=\linewidth]{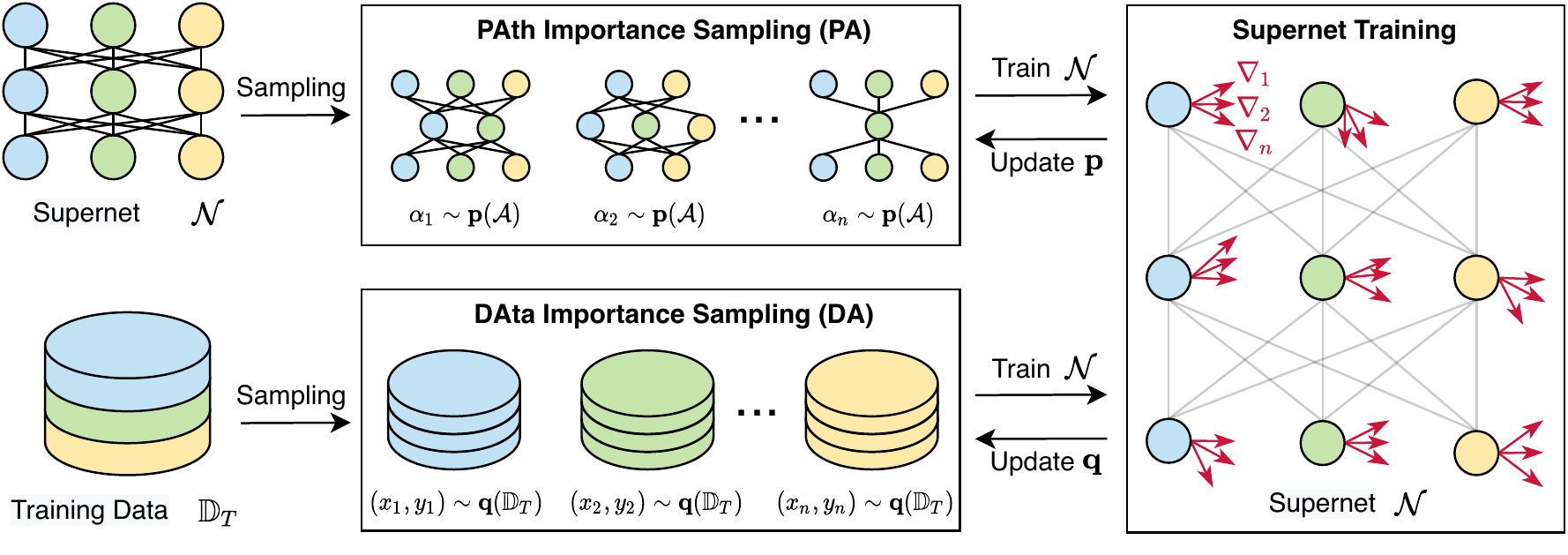}
   \caption{Our supernet training framework includes path importance sampling and data importance sampling. Red arrows show similar gradients of candidate operations via our method. We use the normalized gradient norm to update the path and data sampling distributions.}
   \label{fig:panda_pipeline}
   \vspace{-3mm}
\end{figure*}

\subsection{Path Importance Sampling}

For the sake of simplicity, we first explain how to seek the optimal $\textbf{p}^\star$ with the data sampling distribution $\textbf{q}$ fixed. 
% Please note we conduct alternative optimization of $\textbf{p}$ and $\textbf{q}$ in Algo.\ref{alg:optimization-algorithm}.
Given total training steps $N$, we can re-formulate the expectation in Eq.\ref{eq:supernet_train_obj} as below
\begin{equation}
\mathcal{W}^\star = \arg \min_{\mathcal{W}} \sum_{i=1}^{N} p_i \mathcal{L}(\mathcal{N}(x_i, \alpha_i; \mathcal{W}_{\alpha_{i}}), y_i)
\label{eq:loss_reformulation}
\end{equation}

At $i$-th training step, when we sample a sub-model $\alpha_i$ from the distribution $\textbf{p}(\mathcal{A})$ with the probability $p_i$, the resulting stochastic gradient is given by
\begin{equation}
d_i(p_i)=\frac{1}{Np_i} \nabla_\mathcal{W} \mathcal{L}(\mathcal{N}(x_i, \alpha_i; \mathcal{W}_{\alpha_{i}}), y_i)
\label{eq:path_gradient}
\end{equation}
where the scaling factor $(Np_i)^{-1}$ ensures the gradient $d_i(p_i)$ is an unbiased approximation of the true quantity. 
While in previous works \cite{guo2020single, chu2021fairnas} using the uniform sampling strategy, the sampling probability of $\alpha_i$ is $p_i=\frac{1}{N}$. 
We expect to minimize the gradient variance in Eq.\ref{eq:path_gradient} by optimizing the sampling distribution $\textbf{p}$, which can be formulated as the following optimization problem
\begin{equation}
\min_{\textbf{p}} \mathbb{V}[d(\textbf{p})] = \mathbb{E}\left[d^{\top} d\right]-\mathbb{E}\left[d\right]^{\top} \mathbb{E}\left[d\right]
\label{eq:path_grad_problem}
\end{equation}
By introducing Eq.\ref{eq:path_gradient} into Eq.\ref{eq:path_grad_problem}, we have
\begin{equation}
\begin{aligned}
\mathbb{E}[d^\top d] &= \sum_{i=1}^N p_i \frac{1}{N^2} \frac{1}{p_i^2}\left\|\nabla_\mathcal{W} \mathcal{L}(\mathcal{N}(x_i, \alpha_i; \mathcal{W}_{\alpha_{i}}), y_i)\right\|^2 \\
\mathbb{E}[d] &= \sum_{i=1}^{N}p_id_i =\frac{1}{N} \sum_{i=1}^N \nabla_\mathcal{W} \mathcal{L}(\mathcal{N}(x_i, \alpha_i; \mathcal{W}_{\alpha_{i}}), y_i) \\
\end{aligned}
\end{equation}
We can see that $\mathbb{E}[d]$ is independent of the path sampling distribution $\textbf{p}$, so we can reformulate the problem in Eq.\ref{eq:path_grad_problem} as a constrained optimization problem
\begin{gather*}
\min_{\textbf{p}} \sum_{i=1}^N \frac{1}{N^2} \frac{1}{p_i}\left\|\nabla_\mathcal{W} \mathcal{L}(\mathcal{N}(x_i, \alpha_i; \mathcal{W}_{\alpha_{i}}), y_i)\right\|^2 \\
\mathrm{ s.t. } \sum_{i=1}^N p_i=1 \quad \text{and} \quad p_i \geq 0 \quad \forall i=1,2, \ldots N
\end{gather*}
Since each $\left\|\nabla_\mathcal{W} \mathcal{L}(\mathcal{N}(x_i, \alpha_i; \mathcal{W}_{\alpha_{i}}), y_i)\right\|^2 \geq 0$, the optimal sampling probability $p_i$ must satisfy the inequality constraint and thus we can only consider the equality constraint. 
% \jianchao{Formally, we should have the other multiplier variable for inequality constraints.}
By introducing the Lagrangian multiplier $\lambda$, we have the Lagrangian function $\Psi(\textbf{p}, \lambda)$ as below
\begin{equation}
\begin{aligned}
\Psi(\textbf{p}, \lambda) = & \sum_{i=1}^N \frac{1}{N^2} \frac{1}{p_i}\left\|\nabla_\mathcal{W} \mathcal{L}(\mathcal{N}(x_i, \alpha_i; \mathcal{W}_{\alpha_{i}}), y_i)\right\|^2 \\
& + \lambda(\sum_{i=1}^N p_i-1)
\end{aligned}
\end{equation}
By setting $\frac{\partial \Psi(\textbf{p}, \lambda)}{\partial p_i}=0$, we can get 
\begin{equation}
p_i = \frac{ \|\nabla_\mathcal{W} \mathcal{L}(\mathcal{N}(x_i, \alpha_i; \mathcal{W}_{\alpha_{i}}), y_i)\|}{N\sqrt{\lambda}}
\label{eq:pi_res}
\end{equation}
Applying the equality constraint, we have $\sqrt{\lambda}=\sum_{i=1}^N \frac{\left\|\nabla_\mathcal{W} \mathcal{L}(\mathcal{N}(x_i, \alpha_i; \mathcal{W}_{\alpha_{i}}), y_i)\right\|}{N}$, and further derive the optimal sampling distribution $\textbf{p}^\star$ when
\begin{equation}
p_i^\star = \frac{ \|\nabla_\mathcal{W} \mathcal{L}(\mathcal{N}(x_i, \alpha_i; \mathcal{W}_{\alpha_{i}}), y_i)\|}{\sum_{i=1}^N \|\nabla_\mathcal{W} \mathcal{L}(\mathcal{N}(x_i, \alpha_i; \mathcal{W}_{\alpha_{i}}), y_i)\|}
\label{eq:path_solution}
\end{equation}
Consequently, we can conclude that the optimal path sampling probability $p_i^\star$ is proportional to the normalized gradient norm of the sub-model $\alpha_i$, saying that sampling the sub-model with a larger gradient norm can reduce the gradient variance for the supernet training. 

In practice, we measure the gradient norm of the sub-model $\alpha_i$ as the sum of the gradient norm of its contained candidate operations and use the normalized gradient norm of each candidate operation as their sampling probability. 
We calculate the gradient norm after each conventional backward step and update the sampling probability of candidate operations after each epoch. 
Hence, our optimization for the path sampling distribution $\textbf{p}$ only requires negligible computation and is particularly efficient.

\subsection{Data Importance Sampling}
Now we consider the optimal data sampling distribution $\textbf{q}^\star$ with the path distribution $\textbf{p}$ fixed. 
% Please note we conduct alternative optimization of $\textbf{p}$ and $\textbf{q}$ in Algo.\ref{alg:optimization-algorithm}.
The solution for the $\textbf{q}^\star$ is off-the-shelf as in previous works \cite{needell2014stochastic, zhao2015stochastic, alain2015variance}. 
They have demonstrated that sampling training data according to their normalized gradient norm is helpful to reduce the gradient variance for deep models training, which can be formally expressed as 
\begin{equation}
    q_i^\star \propto \|\nabla_\mathcal{W} \mathcal{L}(\mathcal{N}(x_i, \alpha_i; \mathcal{W}_{\alpha_{i}}), y_i)\|
    \label{eq:data_relationship}
\end{equation}

However, computing the per-sample gradient norm is computationally prohibitive, especially in the context of training deep models, where common deep learning frameworks generally provide the average gradient in a batch-wise manner instead of per-sample-wise. 
Several works \cite{katharopoulos2017biased, katharopoulos2018not, zhang2019autoassist, jiang2019accelerating} have delved into this problem and \cite{katharopoulos2018not} designs an efficient method to approximate the upper bound of the gradient norm for each training data. Specifically, they propose that the gradient of the loss function regarding the pre-activation outputs of the last layer $\nabla_L$ can be deemed an effective estimate of the upper bound, that is 
\begin{equation}
    \sup\{\|\nabla_\mathcal{W} \mathcal{L}(\mathcal{N}(x_i, \alpha_i; \mathcal{W}_{\alpha_{i}}), y_i)\|\} \leq \nabla_L
    \label{eq:upper_bound}
\end{equation}
In this way, we can easily measure the importance of each training data by accessing their upper bound. Take the image classification task as an example, the pre-activation outputs of the last layer $y_L$ are usually followed by a $\mathrm{softmax}$ layer. When using the cross-entropy loss, we can derive the gradient expression for $\nabla_L$ in advance and conveniently compute it during training as below 
\begin{equation}
    \nabla_L = \mathrm{softmax}(y_L) - \mathbbm{1}(y_i)
    \label{eq:g_L_compute}
\end{equation}

The above computation only requires an additional line of code and can be efficiently executed in a mini-batch manner. 
Therefore, we use this approximation to estimate the importance of training data and adopt the normalized results to update the sampling distribution $\textbf{q}$ after each epoch.

\begin{algorithm}[t]
	\caption{Supernet training algorithm of PA\&DA}
	\label{alg:optimization-algorithm}
	{\bfseries Input:} Input training data $\mathbb{D}_T$, supernet $\mathcal{N}$ with weights $\mathcal{W}$, training epochs $n_{epochs}$, training steps $n_{steps}$ per epoch.\\
	{\bfseries Output:} Optimized supernet weights $\mathcal{W}^\star$.
	
	\begin{algorithmic}[1]
		\For{$j=1$ {\bfseries to} $n_{epochs}$}
            \For{$k=1$ {\bfseries to} $n_{steps}$}
                \State Sample a path based on the distribution $\textbf{p}(\mathcal{A})$;
                \State Sample a mini-batch training data based on the distribution $\textbf{q}(\mathbb{D}_T)$;
                \State Train supernet weights $\mathcal{W}$ by gradient descent;
                \State Record gradient norm of the sampled path after back-propagation;
                \State Approximate and record gradient norm of the sampled data using Eq.\ref{eq:g_L_compute}.
            \EndFor
		    \State Linearly increase smoothing parameters $\delta$ and $\tau$;
		    \State Update the path sampling distribution $\textbf{p}(\mathcal{A})$ according to Eq.\ref{eq:path_solution} and add it to uniform distribution;
		    \State Update the data sampling distribution $\textbf{q}(\mathbb{D}_T)$ according to Eq.\ref{eq:data_relationship} and add it to uniform distribution;
		\EndFor
	\end{algorithmic}
\end{algorithm}

\subsection{Importance Sampling NAS}

Our method aims to improve the supernet ranking consistency by reducing the gradient variance during training. We propose a novel and effective importance-based sampling strategy for training the supernet, including path importance sampling and data importance sampling. 
Though we fix one of the sampling distributions in the above derivation, we jointly optimize them in practice. 
We summarize our supernet training algorithm in Algo.\ref{alg:optimization-algorithm}.

\paragraph{PAth Importance Sampling (PA)} 
Following the derived relationship in Eq.\ref{eq:path_solution}, we record and accumulate the gradient norm for each sampled path after the back-propagation. 
We use the normalized gradient norm of each candidate operation as their importance and update the sampling distribution after each epoch. 
To handle those parameter-free operations and avoid the meaningless gradient information at early epochs, we employ a smoothing parameter $\delta$ to add our importance sampling distribution and the uniform sampling distribution. 
We simply linearly increase $\delta$ from 0 to 1 during training and provide a discussion about other changing schemes in our experiments. 

\paragraph{DAta Importance Sampling (DA)}
We use the upper bound in Eq.\ref{eq:g_L_compute} as the importance indicator for our training data. 
After each epoch, we utilize the normalized importance to update the data sampling distribution and adopt an importance sampling strategy with replacement to sample the training indices for the coming epoch. 
Note that if a training instance is not sampled in the current epoch, it will have zero gradient norm in the update, leading to zero sampling probability. 
Analogously, we use a smoothing parameter $\tau$ to add our importance sampling distribution and the uniform sampling distribution together to tackle the above problem.
We linearly increase $\tau$ from 0 to 1 during training and compare other strategies in our ablation studies.

\section{Experiments}
\label{sec:exp}

To demonstrate the effectiveness of PA\&DA in reducing gradient variance during supernet training, we conduct two types of evaluations. 
The first is developed on the NAS-Bench-201 \cite{dong2020bench} using the CIFAR-10 dataset \cite{krizhevsky2009learning} and we provide a comprehensive ranking comparison with other methods. 
The second type is based on the widely-used public search spaces DARTS \cite{liu2019darts} and ProxylessNAS \cite{cai2019proxylessnas}, using CIFAR-10 and ImageNet \cite{krizhevsky2017imagenet} datasets, respectively. 
We conduct an architecture search on these search spaces and compare our search performance with other state-of-the-art methods.
At the end of this section, we further provide extensive ablation studies to analyze our method in depth.

\subsection{Evaluation of Supernet Ranking Consistency}

\paragraph{Search Space}
NAS-Bench-201 \cite{dong2020bench} is a popular NAS benchmark and provides the training and test performance of CIFAR-10, CIFAR-100, and ImageNet-16 for each sub-model in this search space. Sub-models are composed of repeated stacking cells with the same structures. Each cell has four nodes and six edges, and each edge has five candidate operations, leading to $5^6$ architectures in total.

\paragraph{Settings}
We construct the supernet with default settings as NAS-Bench-201 and train it on the CIFAR-10 dataset. 
The smoothing parameters $\delta$ and $\tau$ are linearly increased from 0 to 1. 
We employ total training epochs of 256 with a mini-batch size of 256. 
The SGD optimizer is adopted with an initial learning rate of 0.05, a momentum of 0.9, and a cosine decay strategy. 
After training the supernet, we evaluate the performance of all sub-models on the test dataset by inheriting the pre-trained supernet weights. 

To compare with other methods, we calculate Kendall's Tau (KT) and Precision@Top5\% (P@Top5\%) metrics. 
KT indicates the proportion of correct ranking pairs in all ranking pairs, which measures the supernet overall ranking consistency. 
P@Top5\% is the proportion of predicted top-5\% sub-models in real top-5\% sub-models, showing the ability to identify superb architectures. 

\vspace{-3mm}
\paragraph{Results}
We summarize the results in Tab.\ref{tab:rank_results}. 
The regularization-based or manually-designed methods such as FairNAS, Magic-AT, and SUMNAS not only consume more training time but also perform worse than SPOS. 
NSAS obtains higher KT but lower P@Top5\% and spends nearly an order of magnitude more training time than SPOS. 
Although the splitting methods such as Few-Shot-NAS, GM, and CLOSE achieve better KT, they generally need several times more cost than SPOS. 
In contrast, PA\&DA only requires 0.2 more GPU hours than SPOS and reaches the highest KT and P@Top5\% when compared with others, demonstrating that our training paradigm is effective and beneficial to improving the supernet ranking consistency.

\begin{table}[t]
\small
\begin{center}
\begin{tabular}{lccc}
\toprule
\textbf{Method} & \textbf{Cost} & \textbf{KT}          & \textbf{P@Top5\%}   \\
\midrule
SPOS \cite{guo2020single}           & \textbf{1.6}      & 0.639 $\pm$ 0.030    & 0.211 $\pm$ 0.168 \\
FairNAS$^\dagger$ \cite{chu2021fairnas}       & 5.4     & 0.541 $\pm$ 0.023    & 0.160 $\pm$ 0.034 \\
Magic-AT$^\dagger$ \cite{xu2022analyzing}     & 4.4     & 0.547 $\pm$ 0.059    & 0.019 $\pm$ 0.011 \\
NSAS \cite{zhang2020overcoming}               & 14.6    & 0.653 $\pm$ 0.051    & 0.064 $\pm$ 0.028 \\
SUMNAS$^\dagger$ \cite{ha2021sumnas}          & 22.6    & 0.505 $\pm$ 0.039    & 0.145 $\pm$ 0.061 \\
Few-Shot-25 \cite{zhao2021few}                & 18.6    & 0.696                & -                   \\
GM$^\dagger$-8 \cite{hu2022generalizing}      & 18.0    & 0.656 $\pm$ 0.011    & 0.153 $\pm$ 0.006 \\
CLOSE \cite{zhou2022close}                    & 2.5     & 0.643 $\pm$ 0.050    & 0.031 $\pm$ 0.021 \\
\midrule
PA\&DA                                        & 1.8     & \textbf{0.713 $\pm$ 0.002}    & \textbf{0.301 $\pm$ 0.018} \\
\bottomrule
\end{tabular}
\end{center}
\caption{Ranking results on NAS-Bench-201. Cost: we report the supernet training time in terms of the GPU hours. $^\dagger$: they did not release code thus we implement them following their paper strictly. Few-Shot-25 and GM-8 denote splitting the one-shot supernet into 25 and 8 sub-supernets, respectively.}
\label{tab:rank_results}
\vspace{-4mm}
\end{table}

% \begin{figure}[t]
%   \centering
%   \includegraphics[width=\linewidth]{figure/ipnas_ranking_comp.pdf}
%   \caption{Comparison with other improved methods using the Kendall's Tau and the Precision@top5\%.}
%   \label{fig:ranking_comparison}
% \end{figure}

\subsection{Search Performance on CIFAR-10}

\begin{table*}[th]
	\begin{center}
	\small
		\begin{tabular}{lccccc}
			\toprule
			\multirow{2}*{\textbf{Method}} & \multicolumn{2}{c}{\textbf{Test Accuracy}}& \textbf{Parameters}& \textbf{Search Cost} & \textbf{Search} \\
			\cline{2-3}
			~ & \textbf{Best(\%)} & \textbf{Average(\%)} & \textbf{(M)} & \textbf{(GPU Days)} & \textbf{Method} \\
% 			\textbf{Method} & \textbf{Best Accuracy(\%)} & \textbf{Average Accuracy(\%)} &  \textbf{Parameters(M)}& \textbf{Search Cost (GPU days)}\\
			\midrule
% 			AmoebaNet-A \cite{real2019regularized} & - &  96.66 $\pm$ 0.06 & 3.2 & 3,150 & Evolution \\
			NASNet-A \cite{zoph2018learning} & 97.35 & - &  3.3 & 1,800 & RL\\
			ENAS \cite{pham2018efficient} & 97.11 & - &  4.6 & 0.5 & RL\\
            % \midrule
            % $\triangleright$ \textit{Combined with DARTS} \\
			DARTS \cite{liu2019darts} & - &  97.00 $\pm$ 0.14 & 3.3 & 0.4 & Gradient \\
% 			SNAS \cite{xie2019snas} & - & 97.15 $\pm$ 0.02 & \textbf{2.8} & 1.5 & Gradient\\
            GDAS \cite{dong2019searching} & 97.07 & - &  3.4 & \textbf{0.3} & Gradient\\
            RandomNAS \cite{li2020random} & - & 97.15 $\pm$ 0.08 & 4.3 & 2.7 & Random\\
			% 			EcoNAS \cite{zhou2020econas} & 97.38 $\pm$ 0.02 & 2.9 & 8 \\
            % BayesNAS \cite{zhou2019bayesnas} & - &  97.19 $\pm$ 0.04 & 3.4 & \textbf{0.2} & Gradient \\
% 			R-DARTS(L2) \cite{Zela2020Understanding} & - &  97.05 $\pm$ 0.21 & - & 1.6  & Gradient \\
            % PC-DARTS \cite{xu2019pc} & - &  97.43 $\pm$ 0.07 & 3.6 & 0.1 & Gradient \\
            DARTS-PT \cite{wang2021rethinking} & 97.52 & 97.39 $\pm$ 0.08 & \textbf{3.0} & 0.8 & Gradient \\
            BaLeNAS \cite{zhang2022balenas} & - & 97.50 $\pm$ 0.07 & 3.8 & 0.6 & Gradient \\
            AGNAS \cite{sun2022agnas} & 97.54 & 97.47 $\pm$ 0.003 & 3.6 & 0.4 & Gradient \\
            ZARTS \cite{wang2021zarts} & - & 97.46 $\pm$ 0.07 & 3.7 & 1.0 & Gradient \\
            \midrule
            % $\triangleright$ \textit{Improved one-shot NAS methods} \\
            GDAS-NSAS \cite{zhang2020overcoming} & 97.27 & - & 3.5 & 0.4 & Gradient\\
            RandomNAS-NSAS \cite{zhang2020overcoming} & 97.36 & - & 3.1 & 0.7 & Random\\
            Few-Shot-NAS$^\dagger$ \cite{zhao2021few} & 97.42 & 97.37 $\pm$ 0.06 & 3.8 & 2.8 & Gradient \\
            GM \cite{hu2022generalizing} & 97.60 & 97.51 $\pm$ 0.08 & 3.7 & 2.7 & Gradient \\
            CLOSE \cite{zhou2022close} & - & 97.28 $\pm$ 0.04 & 4.1 & 0.6 & Gradient\\
            \midrule
            PA\&DA & \textbf{97.66} & \textbf{97.52 $\pm$ 0.07} & 3.9 & 0.4 & Random\\
			\bottomrule
		\end{tabular}
	\caption{Comparison with other state-of-the-art methods on the CIFAR-10 dataset using DARTS search space. We report the best and average test accuracy of repeated experiments.$^\dagger$: reported by GM \cite{hu2022generalizing}.}
	\label{tab:darts-results}
	\end{center}
	\vspace{-5mm}
\end{table*}

\paragraph{Search Space}
We use the CIFAR-10 dataset to search for superior cells in the DARTS \cite{liu2019darts} search space. 
The supernet is composed of six normal cells and two reduction cells. 
Normal cells process the feature map without down-sampling, while reduction cells perform down-sampling on the feature map with $\text{stride}=2$ and are located at the $1/3$ and $2/3$ of the total depth of the supernet. 
Each cell consists of seven nodes with four intermediate nodes and fourteen edges with eight candidate operations on each edge. 
We search for the most two powerful operations for each edge to get the final searched cell.

\begin{figure}[t]
  \centering
  \subcaptionbox{Normal Cell}{
  \includegraphics[width=0.48\linewidth]{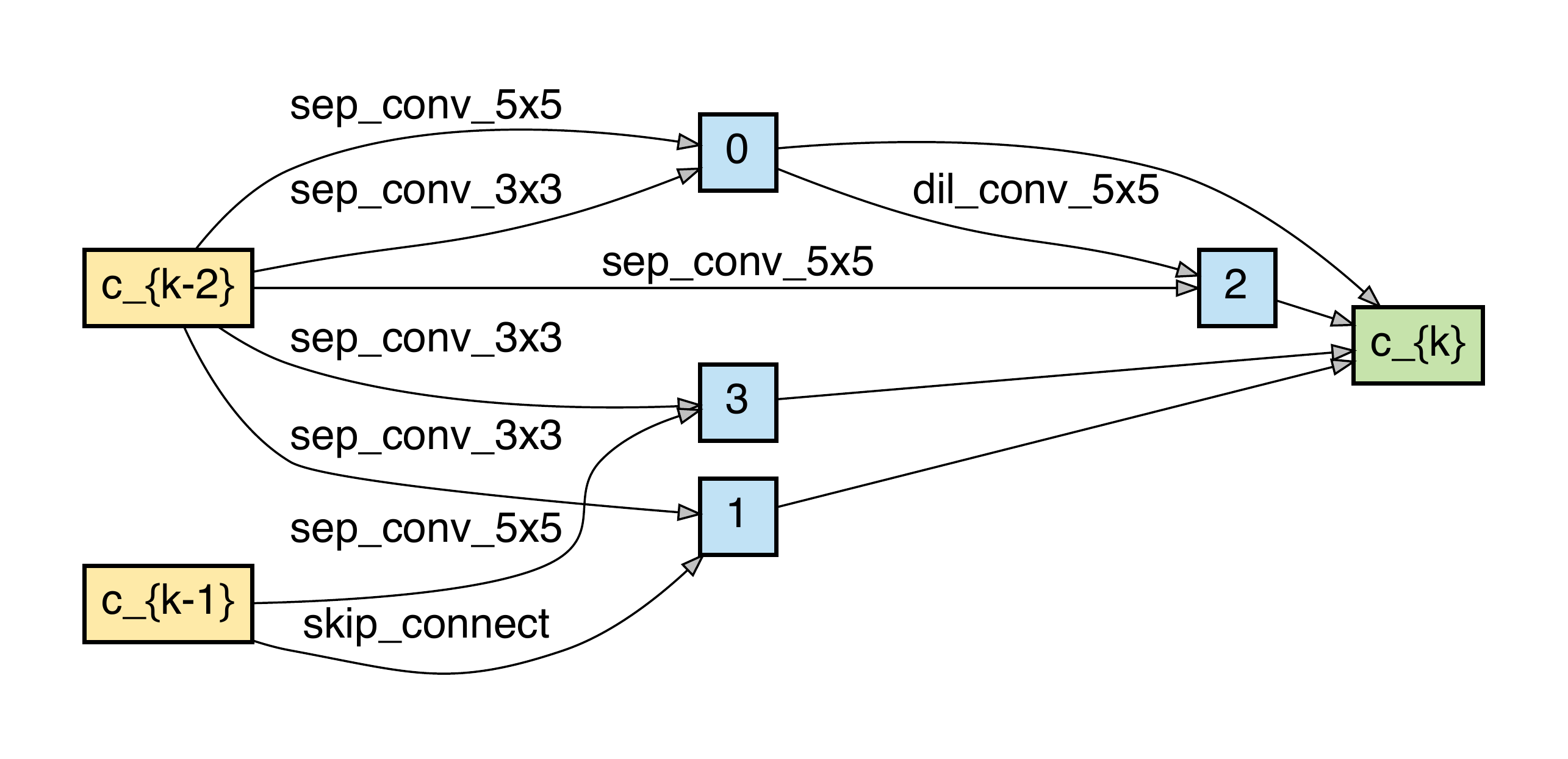}}
  \subcaptionbox{Reduction Cell}{
  \includegraphics[width=0.48\linewidth]{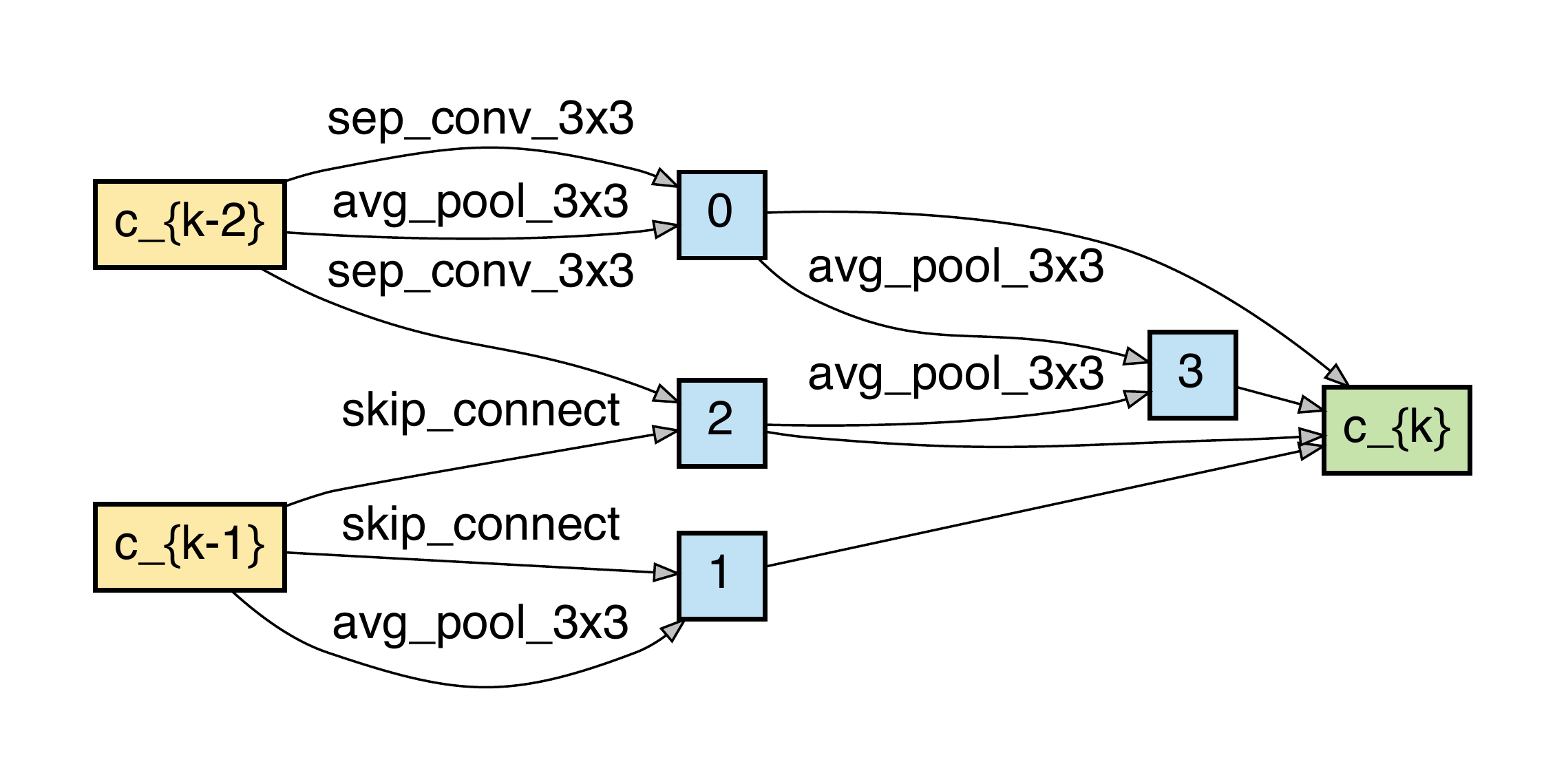}}
  \caption{Our best searched cells in the DARTS search space.}
  \label{fig:best_cell}
  \vspace{-3mm}
\end{figure}

\vspace{-3mm}
\paragraph{Settings}
We follow the settings in NSAS \cite{zhang2020overcoming} to combine our method with RandomNAS \cite{li2020random}. We use the SGD optimizer with momentum 0.9 and weight decay 3e-4 to train the supernet for 50 epochs. The initial learning rate is 0.025 and is then decayed to 0.001 by a cosine strategy. After the supernet training, we randomly search for 60 rounds and evaluate 100 sub-models at each round to select the most promising architecture. 
By re-training the searched architecture, we compare the top-1 classification accuracy with other methods. 
We visualize the best searched cell in Fig.\ref{fig:best_cell}, and provide other cells and re-training details in our Supp.

\paragraph{Results}
We report the best and average test accuracy from repeated experiments with three random seeds in Tab.\ref{tab:darts-results}. 
As can be seen, our method achieves the highest average test accuracy 97.52 $\pm$ 0.07, surpassing the original DARTS and its advanced variants.
When compared with other improved one-shot NAS methods such as NSAS, Few-Shot-NAS, GM, and CLOSE, our method consistently outperforms them with the least search cost. 

Our best cells are shown in Fig.\ref{fig:best_cell}. 
We can observe that both the normal cell and the reduction cell have the $skip\_connect$ operation from the input nodes, leading to a residual link with other operations. 
As pointed out in \cite{wan2022redundancy}, such a ResNet-style residual link is helpful for achieving state-of-the-art performance, demonstrating that our method excels in identifying excellent architectures.

\subsection{Search Performance on ImageNet}

\paragraph{Search Space}
We use the chain-like search space as proposed in ProxylessNAS \cite{cai2019proxylessnas}, including 21 searchable layers in the supernet.  
We search for lightweight MobileNet \cite{sandler2018mobilenetv2} blocks by exploring the kernel sizes \{$3, 5, 7$\} and expansion rates \{$3, 6$\} for the searchable blocks. 
A searchable $skip\_connect$ is added for the blocks without down-sampling, leading to 7 or 6 candidate operations per layer.

\paragraph{Settings}
We utilize our method to train the supernet on 8 GPU cards for 120 epochs, with a total batch size of 2048. 
SGD optimizer is adopted with the weight decay 4e-5 and momentum 0.9. 
The initial learning rate is 0.5 and is decayed to 5e-4 by a cosine strategy. 
After training the supernet, we use an evolutionary search algorithm to search for top-performing architectures with the FLOPs constraint 400 M. 
The evolutionary search lasts for 20 epochs in total. At each epoch, we maintain a population with 50 sub-models, including 25 sub-models from mutation and crossover respectively. 
We retrain our searched architecture on the ImageNet training dataset and evaluate its performance on the validation dataset. 
The detailed retraining configuration and our searched architecture are provided in our Supp.

\paragraph{Results}
We report the results in Tab.\ref{tab:mobilenet-results}. 
Our PA\&DA surpasses DA-NAS, FairNAS-A, and SUMNAS-M with a bit more FLOPs. When compared with SPOS, ProxylessNAS, MAGIC-AT, Few-Shot NAS, and GM, our searched architecture is smaller and obtains the highest top-1 accuracy 77.3, suffice to demonstrate the efficacy of our method.

\begin{table}[t]
\centering
\small
\begin{tabular}{lcccc}
\toprule
\multirow{2}*{\textbf{Method}} & \textbf{Params.} & \textbf{FLOPs} & \textbf{Top-1} & \textbf{Top-5} \\
~ & \textbf{(M)} & \textbf{(M)} & \textbf{(\%)} & \textbf{(\%)} \\
\midrule
NASNet-A \cite{zoph2018learning} & 5.3 & 564 & 74.0 & 91.3 \\
AmoebaNet-A \cite{real2019regularized} & 5.1 & 555 & 74.5 & 92.0 \\
MnasNet-A1 \cite{tan2019mnasnet} & \textbf{3.9} & \textbf{312} & 75.2 & 92.5 \\
PNAS \cite{liu2018progressive} & 5.1 & 588 & 74.2 & 91.9 \\
\midrule
DA-NAS \cite{dai2020data} & - & 389 & 74.6 & -  \\
SPOS \cite{guo2020single} & 5.4 & 472 & 74.8 & - \\
FBNet-C \cite{wu2019fbnet} & 5.5 & 375 & 74.9 & - \\
ProxylessNAS \cite{cai2019proxylessnas} & 7.1 & 465 & 75.1 & 92.3 \\
FairNAS-A \cite{chu2021fairnas} & 4.6 & 388 & 75.3 & - \\
MAGIC-AT \cite{xu2022analyzing} & 6.0 & 598 & 76.8 & 93.3 \\ 
SUMNAS-M \cite{ha2021sumnas} & 5.1 & 392 & 77.1 & - \\
Few-Shot NAS \cite{zhao2021few} & 4.9 & 521 & 75.9 & - \\
GM \cite{hu2022generalizing} & 4.9 & 530 & 76.6 & 93.0 \\
\midrule
PA\&DA & 5.3 & 399 & \textbf{77.3} & \textbf{93.5}\\
\bottomrule
\end{tabular}
\caption{Comparison with other state-of-the-art methods on the ImageNet dataset using the ProxylessNAS search space.}
\label{tab:mobilenet-results}
\vspace{-5mm}
\end{table}

\subsection{Ablation Studies}

\begin{figure*}[t]
  \centering
  \subcaptionbox{CIFAR-10}{
  \includegraphics[width=0.24\linewidth]{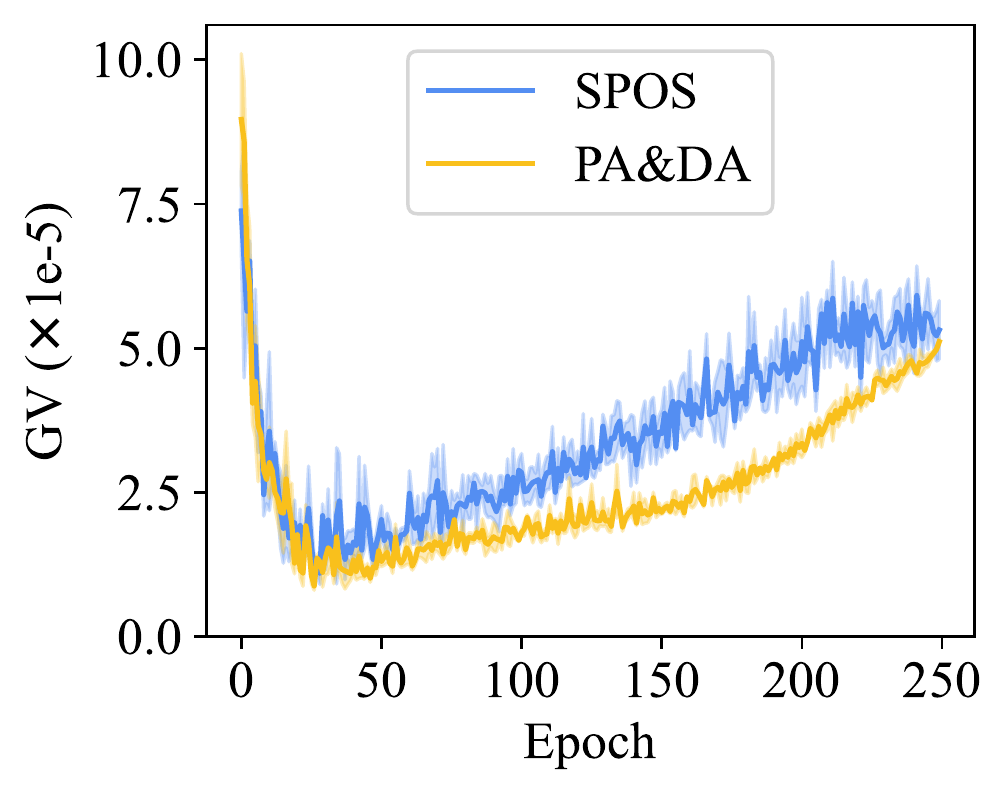}}
  \subcaptionbox{CIFAR-100}{
  \includegraphics[width=0.24\linewidth]{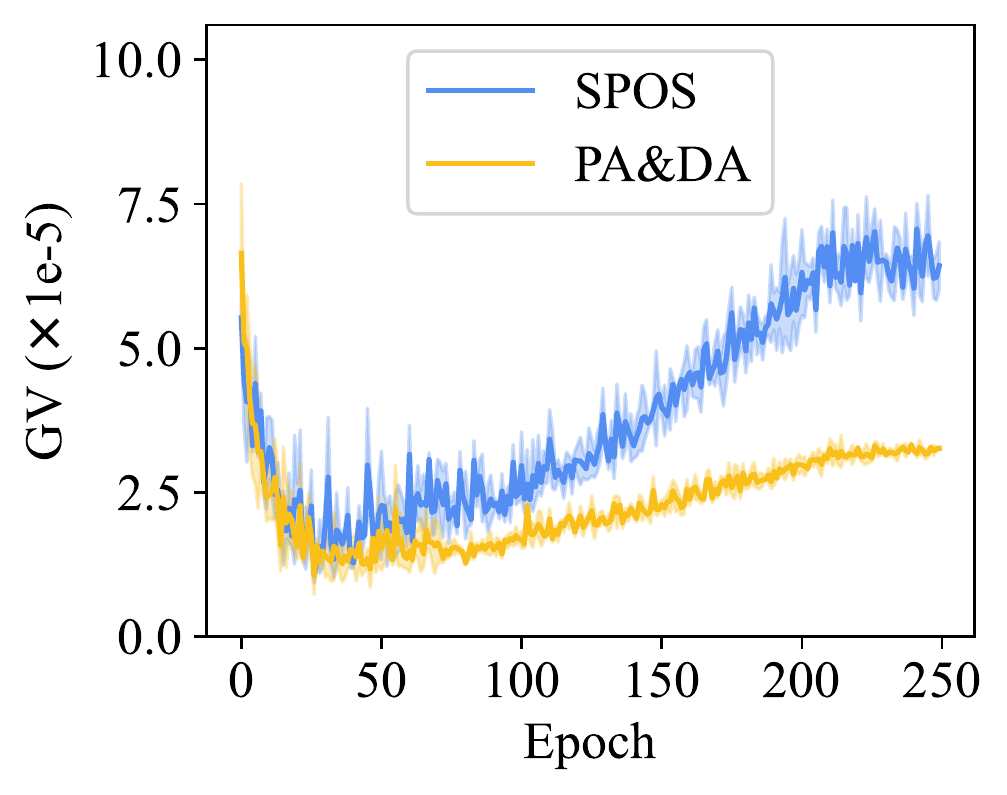}}
  \subcaptionbox{ImageNet16-120}{
  \includegraphics[width=0.24\linewidth]{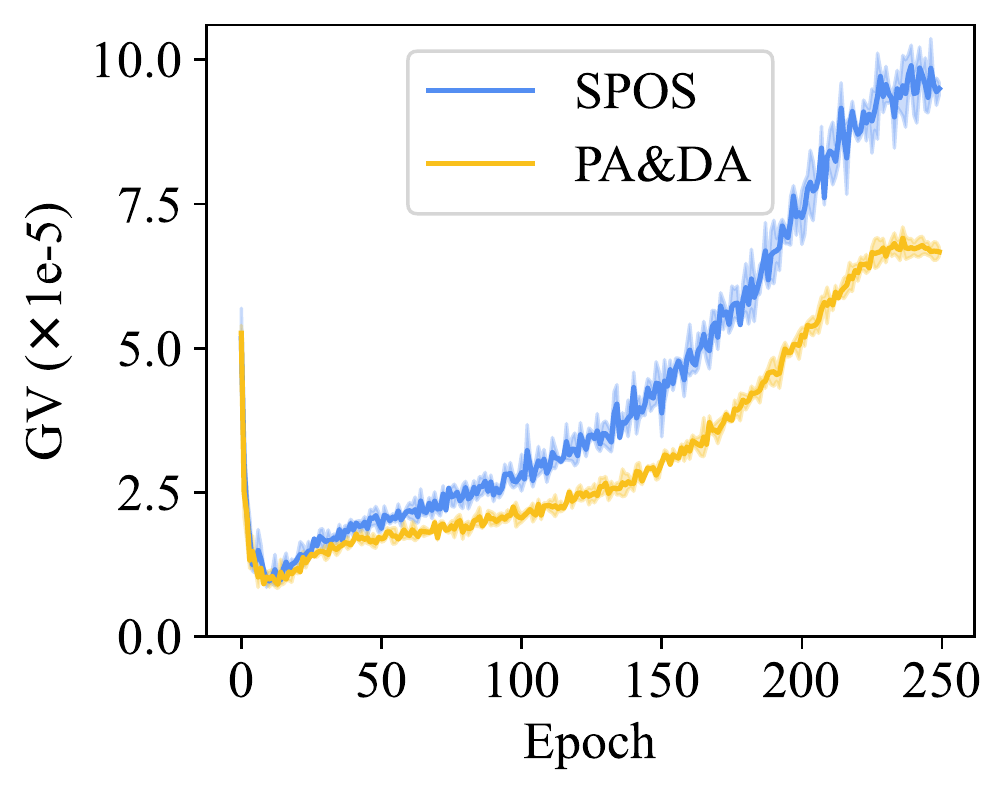}}
  \subcaptionbox{Ranking Comparison}{
  \includegraphics[width=0.24\linewidth]{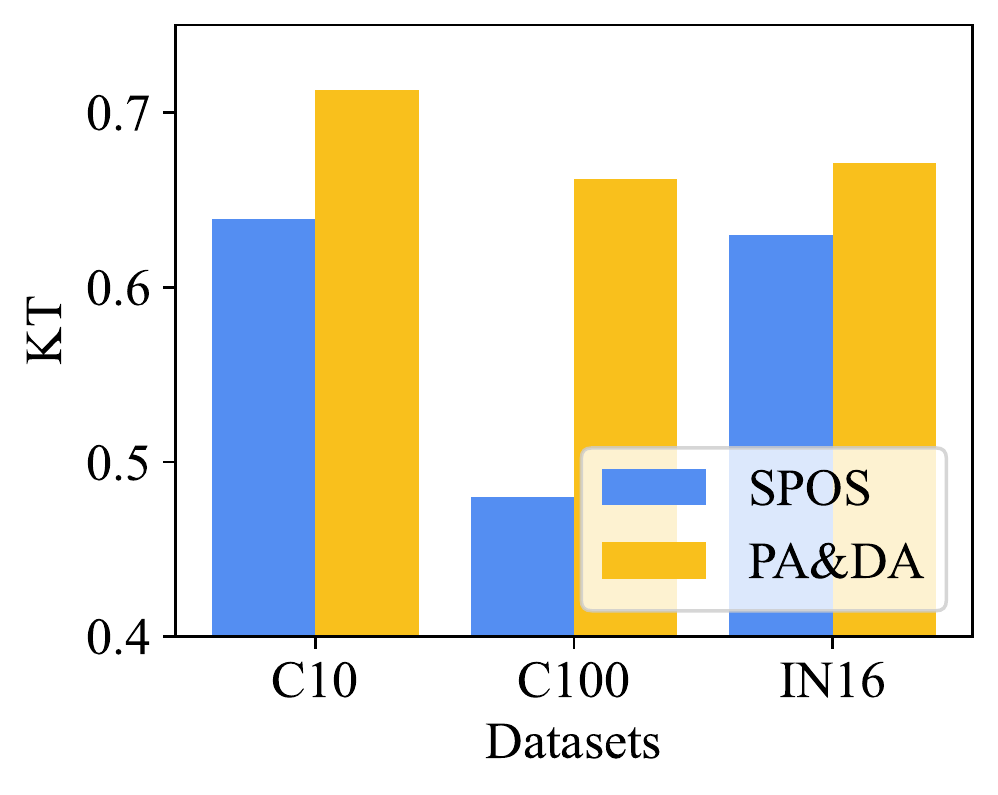}}
  \caption{KT and GV on NAS-Bench-201 using three datasets. C10: CIFAR-10, C100: CIFAR-100, IN16: ImageNet16-120.}
  \label{fig:kt_gv_multi_datasets}
  \vspace{-3mm}
\end{figure*}

% \paragraph{Predicted sub-models distribution}
% To prove that the low-gradient-variance training condition is beneficial, we further plot the distribution of the number of sub-models with respect to the predicted accuracy from one experiment in the right of Fig.\ref{fig:grad_var_arch_dist}. 
% We can see that sub-models predicted by the supernet from PA\&DA distribute more evenly rather than gather around as SPOS, and top-performing architectures attain higher predicted accuracy. 
% These phenomena indicate that PA\&DA distinguishes sub-models clearer and thus achieves a better ranking result 0.713 $\pm$ 0.002 than SPOS 0.639 $\pm$ 0.030, demonstrating that PA\&DA cultivates a better-generalized supernet.

% \begin{figure}[t]
%   \centering
%   \includegraphics[width=\linewidth]{figure/ipnas_grad_var_comp.pdf}
%   \caption{Left: Comparison of the gradient variance between our method and the baseline. Right: The distribution of the number of sub-models with respect to the predicted accuracy.}
%   \label{fig:grad_var_arch_dist}
% \end{figure}

\begin{figure}[t]
  \centering
  \subcaptionbox{KT and GV w.r.t Batch Size}{
  \includegraphics[width=0.48\linewidth]{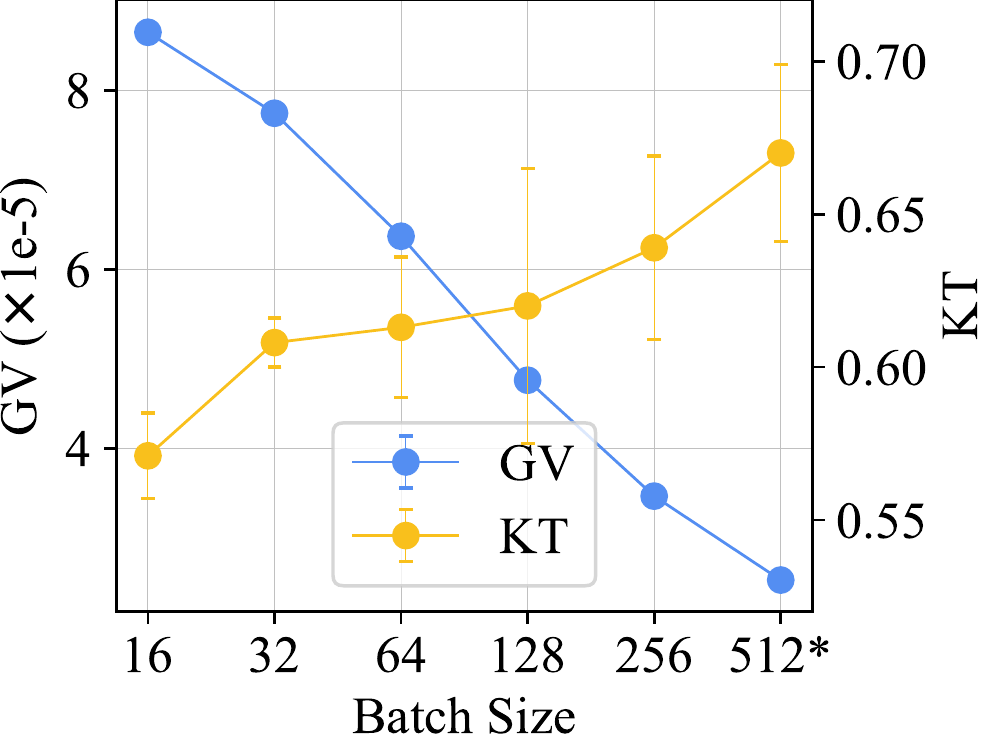}}
  \subcaptionbox{Trainability comparison}{
  \includegraphics[width=0.48\linewidth]{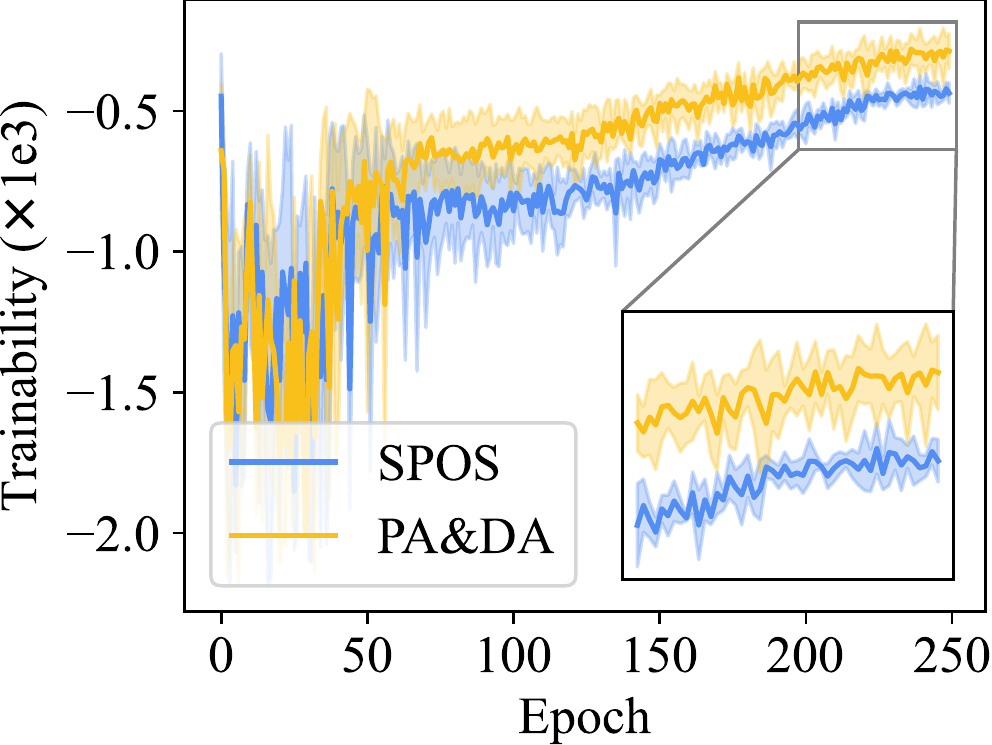}}
  \caption{Effect of various batch sizes and trainability comparison.}
  \label{fig:ablation_batch_trainability}
  \vspace{-3mm} 
\end{figure}

\paragraph{Effect of batch size}
It is common that larger batch sizes (BS) can stabilize the training of deep models with lower gradient variance. 
We conduct an ablation study using the SPOS \cite{guo2020single} by varying BS from 16 to 512 to validate this phenomenon during supernet training. 
As larger BS tends to have fewer training steps, we use more steps for BS 512 to ensure sufficient convergence. 
As shown in Fig.\ref{fig:ablation_batch_trainability}(a), we can observe that GV decreases and KT increases monotonically as BS becomes larger, and BS 512 obtains the best KT 0.670 $\pm$ 0.029. 
These results further verify that lower gradient variance benefits the supernet training, which exactly meets our idea. 
However, PA\&DA does not require more training steps than BS 512 and gets higher KT 0.713 $\pm$ 0.002, which is much more efficient and effective.
\vspace{-3mm}

\paragraph{Effect of schedules for smoothing parameters}
To pre-load data indices for efficiency, we update the sampling probability of DA after each epoch. We explore two changing styles for $\tau$: linearly decrease and increase, and evaluate the distribution granularity sample-wise or class-wise in Tab.\ref{tab:ablation-smoothing}. 
Notice that using a sample-wise distribution and linearly increasing $\tau$ yields the best result. 
As for PA, we investigate the update frequency for the sampling distribution and also two changing styles for $\delta$. Results show that updating the sampling probability per epoch and linearly increasing $\delta$ is better. 
Both results suggest a linearly increase schedule for smoothing parameters, showing that the importance sampling is preferable in the late of training.
\vspace{-1mm}

\begin{table}[ht]
\centering
\begin{tabular}{ccccc}
\toprule
\textbf{Module} & \textbf{Freq.} & \textbf{Style} & \textbf{Gran.} & \textbf{KT} \\
\midrule
\multirow{3}*{DA} & PE  & $\downarrow$ & Instance & 0.643 $\pm$ 0.021 \\
~ & PE & $\uparrow$ & Class & 0.637 $\pm$ 0.024 \\
~ & PE & $\uparrow$ & Instance & \textbf{0.644 $\pm$ 0.014} \\
\midrule
\multirow{3}*{PA} & PS & $\downarrow$ & Path & 0.663 $\pm$ 0.008 \\
~ & PE & $\downarrow$ & Path & 0.667 $\pm$ 0.003 \\
~ & PE & $\uparrow$ & Path & \textbf{0.699 $\pm$ 0.004} \\
\bottomrule
\end{tabular}
\caption{Ranking performance w.r.t the smoothing parameters and update schedules for DA and PA. PE: update the sampling distribution per epoch, PS: update the sampling distribution per step.}
\label{tab:ablation-smoothing}
\vspace{-3mm}
\end{table}

\paragraph{Effect of DA and PA}
We conduct the ablation study for DA and PA in Tab.\ref{tab:ablation_da_pa}. 
When DA and PA are both disabled, our method degenerates to the baseline method SPOS \cite{guo2020single}. 
When either one is applied, we can obtain higher KT and P@Top5\%. 
Furthermore, both modules cooperate well with each other, and using them together yields the best result. 
Besides, we empirically observe that PA contributes more performance gains than DA. 
% \vspace{-3mm}

% \begin{wrapfigure}{r}{0.47\linewidth}
%   \vspace{-5mm}
%   \centering
%   \includegraphics[width=\linewidth]{figure/panda_ablation_errorbar.pdf}
%   \caption{Ablation study.}
%   \label{fig:panda_ablation_errbar}
%   \vspace{-5mm}
% \end{wrapfigure}

\begin{table}[ht]
\centering
\begin{tabular}{cccc}
\toprule
\textbf{DA}          & \textbf{PA}          & \textbf{KT} & \textbf{P@Top5\%} \\
\midrule
-            & -            & 0.639 $\pm$ 0.030 & 0.211 $\pm$ 0.168 \\
$\checkmark$ & -            & 0.644 $\pm$ 0.014 & 0.225 $\pm$ 0.049 \\
-            & $\checkmark$ & 0.699 $\pm$ 0.004 & 0.299 $\pm$ 0.008 \\
$\checkmark$ & $\checkmark$ & \textbf{0.713 $\pm$ 0.002} & \textbf{0.301 $\pm$ 0.018} \\
\bottomrule
\end{tabular}
\caption{Ablation study for PA and DA.}
\label{tab:ablation_da_pa}
% \vspace{-5mm}
\end{table}

\section{Analysis and Discussions}

\subsection{Gradient Variance Comparison}
To exhibit the benefit of PA\&DA for reducing the gradient variance during supernet training, we further compare PA\&DA with the baseline method SPOS \cite{guo2020single} on CIFAR-10, CIFAR-100, and ImageNet-16-120 datasets using the NAS-Bench-201 \cite{dong2020bench} search space. 
We repeat the experiments with three different seeds and record the supernet gradient variance of each epoch in Fig.\ref{fig:kt_gv_multi_datasets}.
As the training goes on and smoothing parameters linearly increases, the original uniform sampling strategy for path and data gradually shifts to biased sampling, making PA\&DA achieve a lower gradient variance than the baseline. 
Due to this advantage, the supernet trained by PA\&DA obtains higher ranking consistency in three datasets as shown in Fig.\ref{fig:kt_gv_multi_datasets}(d).

\subsection{Trainability of Paths from PA}
As analyzed above, PA plays a more critical role than DA. 
To explore its key advantage, we adopt the trainability from TE-NAS \cite{chen2021neural} to measure the sampled path at each training step. 
Results are shown in Fig.\ref{fig:ablation_batch_trainability}(b). 
We can see that the sampled path of PA constantly enjoys a higher trainability than the baseline, especially at the end of the training, which explains the faster convergence and better generalization performance from PA.

\section{Conclusion}
\label{sec:conclusion}
In this work, we reduce the gradient variance for the supernet training by jointly optimizing the path and data sampling distributions to improve the supernet ranking consistency. 
We derive the relationship between the gradient variance and the sampling distributions and use the normalized gradient norm to update both distributions. 
Extensive experiments demonstrate the effectiveness of our method.
In the future, we will further explore more effective methods to reduce gradient variance for supernet training.

\section*{Acknowledgments}
This work was supported in part by the National Key R\&D Program of China under grant No. 2018AAA0102701 and in part by the National Natural Science Foundation of China under Grant No. 62176250 and No.62203424.

%\clearpage

%%%%%%%%% REFERENCES
{\small
\bibliographystyle{ieee_fullname}
\bibliography{egbib}
}

%%%%%%%%% Appendix
\clearpage
\onecolumn
\appendix
\setcounter{table}{4}
\setcounter{figure}{5}
\setcounter{equation}{13}

% \section{Overview}

\section{Calculation of the supernet gradient variance}
\label{sec:var_cal}
During the supernet training, we record the gradient $d_w$ of each parameter $w \in \mathcal{W}$ after each training step, using the gradient generated by the normal back-propagation. 
After an epoch of training, we utilize the recorded information to compute the gradient variance $\sigma_w^2$ of each parameter $w \in \mathcal{W}$ as below, 
\begin{equation}
    \sigma_w^2 = \frac{1}{S} \sum_{s=1}^S (d_{w_s} -  \mu_{w})
\end{equation}
where $S$ is the sampled times for the parameter $w$ and $\mu_{w}$ stands for the average gradient of $w$ during updates. 

By collecting the gradient variance of each parameter $w \in \mathcal{W}$, we further calculate the average value to represent the supernet gradient variance $\sigma_\mathcal{N}^2$, which can be formulated as 
\begin{equation}
\sigma_\mathcal{N}^2 = \mathbb{E}_{w \in \mathcal{W}}[\sigma_w^2]
\label{eq:supernet_gd_var}
\end{equation}

\section{Re-training configuration}
\label{sec:train_config}

\subsection{Settings in the DARTS search space}
The experimental settings are consistent with previous works \cite{liu2019darts, wang2021rethinking} to ensure a fair comparison. 
By stacking the searched normal and reduction cells, the final architecture is consisted of 20 layers and 36 channels. 
The final architecture is re-trained on a single GPU \footnote{All of our experiments were conducted on the NVIDIA Tesla V100 GPU.\label{fnt: gpu_type}} by a total of 600 epochs using the training dataset and evaluated on the test dataset to get the top-1 accuracy. 
The initial learning rate is 2.5e-2 and is then decayed to zero via a cosine strategy. 
We use the SGD optimizer with the weight decay 3e-4, momentum 0.9, and the training batch size 96. 
The auxiliary head with a weight of 0.4 and the drop path \cite{larsson2016fractalnet} with a probability of 0.2 are both adopted to mitigate over-fitting. 
We use the Cutout \cite{devries2017improved} technique with the length 16 to augment the training data. 
Besides, we set the threshold of the gradient norm clipping as 5 for all trainable parameters.

\subsection{Settings in the ProxylessNAS search space}
The final architecture contains 21 layers, one of which is the Identity layer. 
We use 8 GPUs \textsuperscript{\ref{fnt: gpu_type}} in parallel to re-train our searched architecture on the ImageNet training dataset for 450 epochs and evaluate its performance on the validation dataset. 
We use the RMSpropTF optimizer with an initial learning rate of 0.16 and a step decay scheduler, which decays the learning rate per 2.4 epochs with a reduction rate of 0.97. 
The weight decay is 1e-5 and the momentum is 0.9. 
To mitigate over-fitting, we adopt the AutoAug \cite{cubuk2019autoaugment} and RE \cite{zhong2020random} for the data augmentation, and utilize both the drop path \cite{larsson2016fractalnet} and Dropout \cite{srivastava2014dropout} with the same rate 0.2. 
At the initial stage of the training, we utilize a small learning rate of 1e-6 for warm-up by 3 epochs. 
During training, the moving average technique is employed to smooth the model weights with a rate of 0.9999.
\section{Visualization}

\subsection{Searched cells in DARTS search space}

\begin{figure*}[h]
  \centering
  \subcaptionbox{PA\&DA-1 (normal cell)}{
  \includegraphics[width=0.48\linewidth]{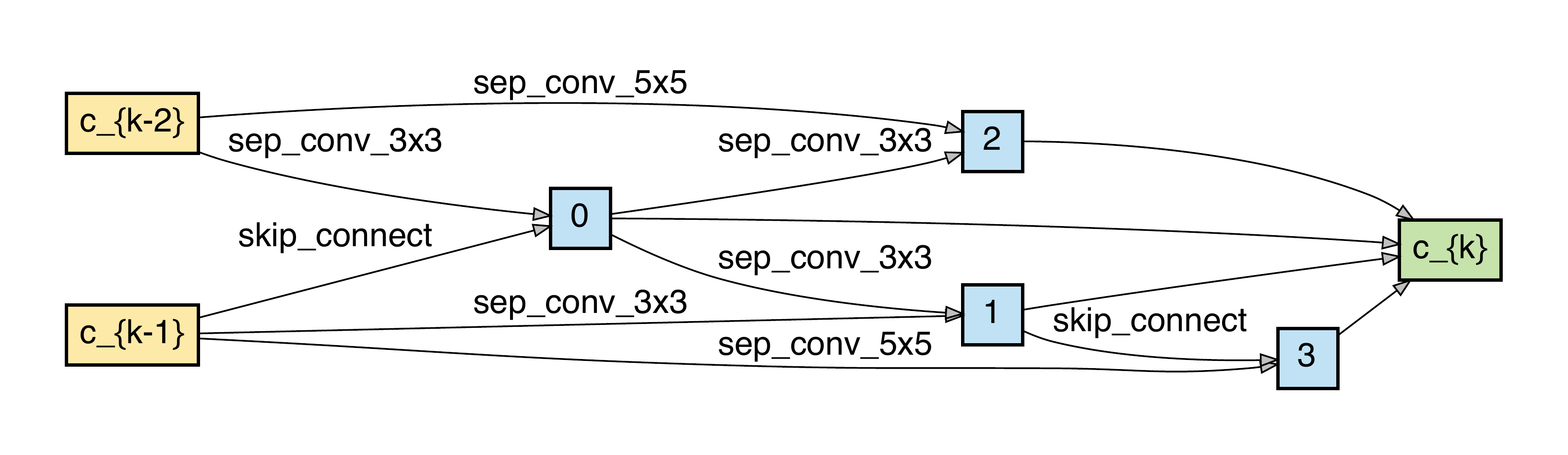}}
  \subcaptionbox{PA\&DA-1 (reduction cell)}{
  \includegraphics[width=0.48\linewidth]{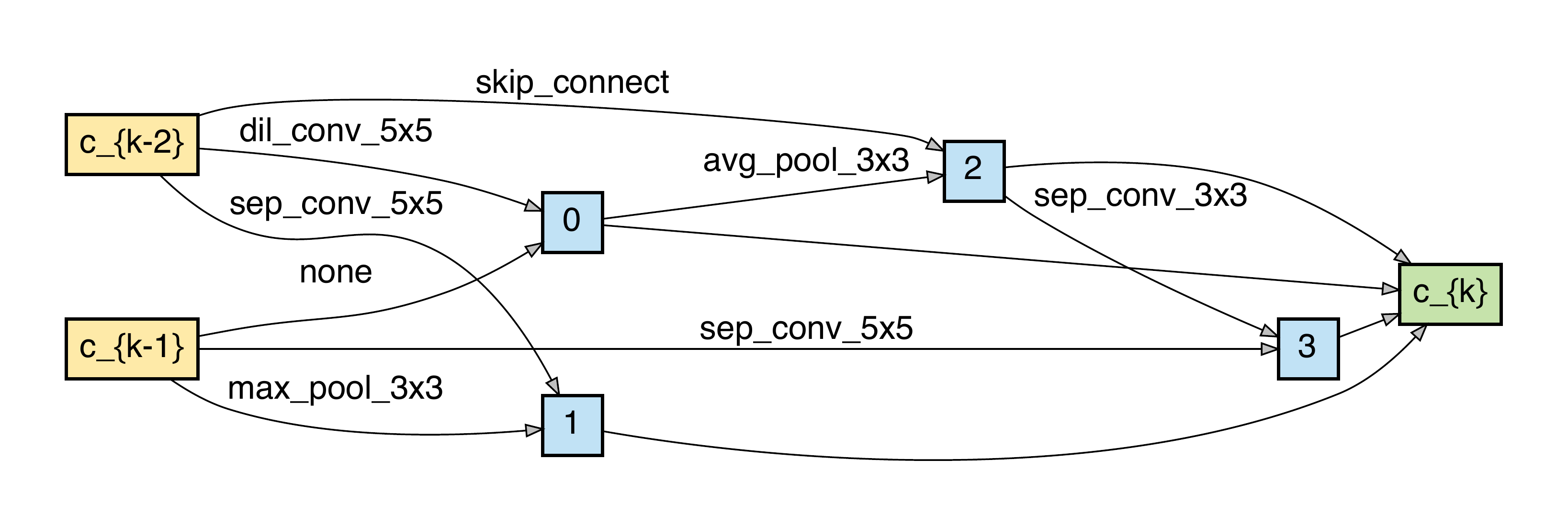}}
  \subcaptionbox{PA\&DA-2 (normal cell)}{
  \includegraphics[width=0.48\linewidth]{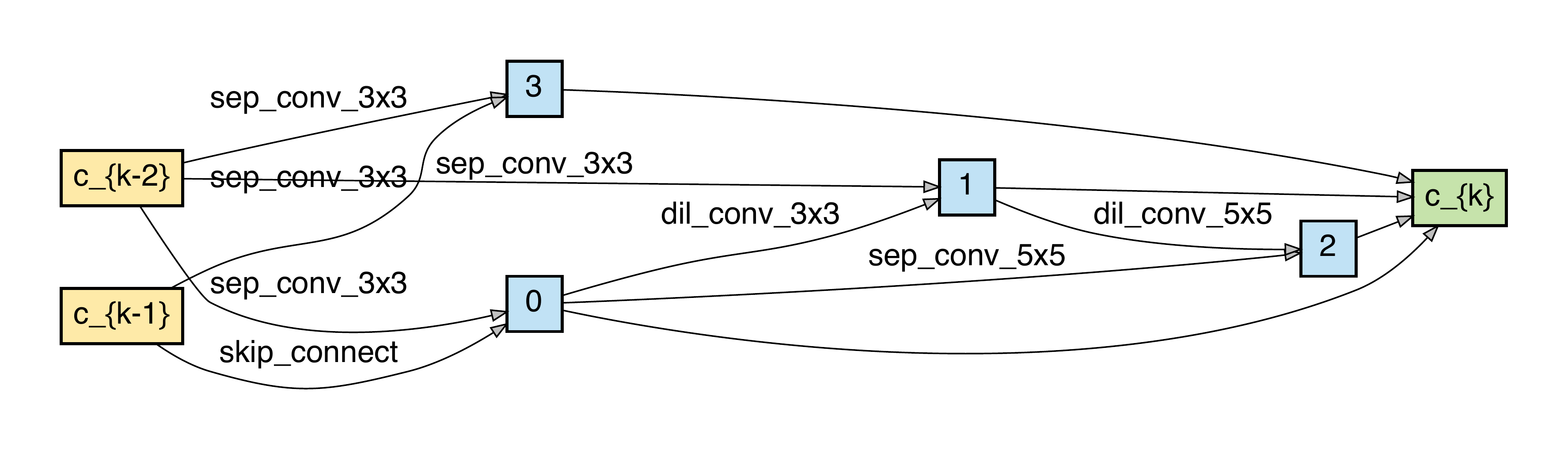}}
  \subcaptionbox{PA\&DA-2 (reduction cell)}{
  \includegraphics[width=0.48\linewidth]{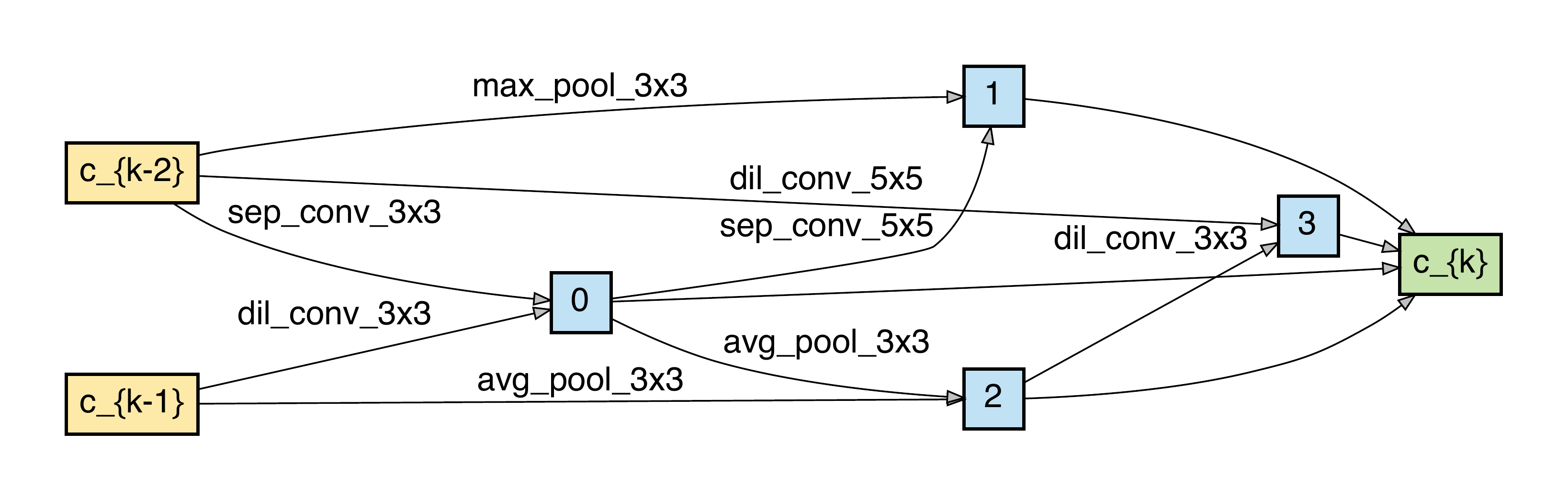}}
  \caption{Our searched cells in the DARTS search space.}
  \label{fig:other_darts_cells}
\end{figure*}

Our best-searched cells have been shown in Fig.3 of our main text and we present the other two searched cells in Fig.\ref{fig:other_darts_cells}. 
Although these searched cells have different operations and typologies, we can find a common characteristic of them: all the searched normal cells have many $sep\_conv\_3{\times}3$ operations and have one $skip\_connect$ operation from the input node to one of the intermediate nodes. 
We conjecture that these merits lead to the superior performance of the searched cells.

\clearpage

\subsection{Searched architectures in ProxylessNAS search space}

\begin{figure}[H]
  \centering
  \includegraphics[width=\linewidth]{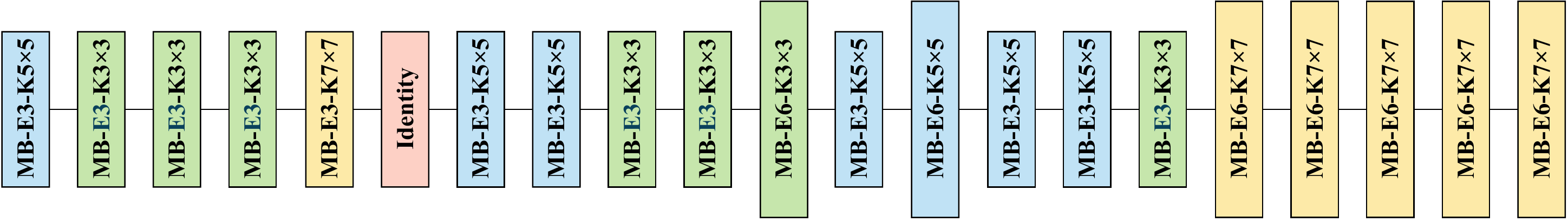}
  \caption{Our searched architecture in the ProxylessNAS search space. The expansion rate of short squares is 3, while being 6 for long squares. Colors of green, blue, and yellow denotes the kernel size $3{\times}3, 5{\times}5$ and $7{\times}7$ of the depth-wise convolution in the MobileNet block, respectively. The red square stands for the layer with the Identity operation.}
  \label{fig:best_arch_mobilenet}
\end{figure}

As shown in Fig.\ref{fig:best_arch_mobilenet}, slimmer channels and smaller receptive fields are preferable at the beginning of the network, thus our searched architecture adopts smaller expansion rates and kernel sizes in shallow layers. 
On the contrary, the last 5 layers all choose the expansion rate 6 with the largest kernel size 7, demonstrating that more channels and larger receptive fields are necessary for encoding the semantic information.

\end{document}